# Can Synthetic Data Improve Object Detection Results for Remote Sensing Images?

Weixing Liu, Jun Liu and Bin Luo

*Abstract*—Deep learning approaches require enough training samples to perform well, but it is a challenge to collect enough real training data and label them manually. In this letter, we propose the use of realistic synthetic data with a wide distribution to improve the performance of remote sensing image aircraft detection. Specifically, to increase the variability of synthetic data, we randomly set the parameters during rendering, such as the size of the instance and the class of background images. In order to make the synthetic images more realistic, we then refine the synthetic images at the pixel level using CycleGAN with real unlabeled images. We also fine-tune the model with a small amount of real data, to obtain a higher accuracy. Experiments on NWPU VHR-10, UCAS-AOD and DIOR datasets demonstrate that the proposed method can be applied for augmenting insufficient real data.

*Index Terms*—Object detection, synthetic images, domain randomization, image translation.

## I. INTRODUCTION

Deep convolutional neural networks (DCNNs) can achieve excellent performances in object detection, in which large-scale labeled training data play a crucial role [1]. As a special application field, remote sensing object detection involves a large amount of data, but with a lack of annotation information. The datasets [2]–[4] that are commonly used in remote sensing for object detection are mostly labeled manually, which is time-consuming and costly [5].

Synthetic data have been widely employed in several fields of deep learning. Examples of such synthetic datasets are [6], [7]. However, there are few synthetic data in the field of remote sensing object detection. In recent years, it has become possible to acquire crowdsourced 3D models online, which is an attractive way to generate an unlimited volume of training data. By rendering 3D models of a virtual scene, we can automatically generate synthetic images with labels for training. However, to build a realistic virtual scene, we need not only high-fidelity texture, but also a complete scene model. Hence, several studies have investigated using real images as the virtual scene background, which can improve the performance when the volume of real training data is limited [8], [9]. Generally speaking, synthetic images are lacking in texture and context, while real images are rich in visual details [8], [10]. Because of this so-called "reality gap" [11], synthetic datasets are mainly used to expand existing real datasets, rather than being used as training data on their own.

In this study, we attempted to use synthetic data to improve the results of aircraft detection in remote sensing images. We argue that: 1) the content of the synthetic imagery should be varied so that the data distribution is wide enough to cover the real distribution; and 2) the style of the synthetic imagery should be as realistic as possible, to reduce the domain shift. To improve the performance on real tasks with the help of synthetic data, the typical method including pixel-level, feature-level and model-level domain adaptation [12]. As we don't want to modify the structure of general object detectors, we focus on pixel-level adaptation and fine-tuning. Similar ideas put forward in the remote sensing field include the work of Yan *et al* [13]; however, they ignored the reality gap.

To increase the variability of synthetic data, domain randomization (DR) [14] has become a popular way to create a variety of simulated environments with randomized properties (e.g., color, texture, lighting). If the virtual environment is diverse enough, a model trained in the simulated environment can be extended to the real world, without requiring real data. Xia *et al* pointed out that the difficulties of remote sensing image object detection include the scale variables of object instances, complex background [15]. Therefore, we generate aircraft instances with large scale changes based on the idea of domain randomization, with a variety of typical scene class image as the background.

To reduce the gap between synthetic and real data, pixel-level adversarial adaptation attempts to transfer synthetic domain images to real domain images [16], [17]. The idea of using a conditional generative adversarial network (cGAN) for image-to-image mapping was first proposed by Isola *et al*. [16], who released the pix2pix method. Nevertheless, collecting paired samples for this approach is difficult in most cases. To address this issue, researchers [17] have proposed to translate unpaired data based on cycle consistency. In the proposed approach, we use CycleGAN to refine the synthetic data while preserving the semantic layout.

In summary, we extended the domain randomization approach by refining the synthetic images with a generative adversarial network (GAN)-based image translator. Our contributions are as follows:

1) We propose a general system for automatically generating content-rich, realistic synthetic images with ground-truth annotations. To cover the distribution of real data, we vary the scale of synthetic airplane instances, and choose various scenes class images as the background.

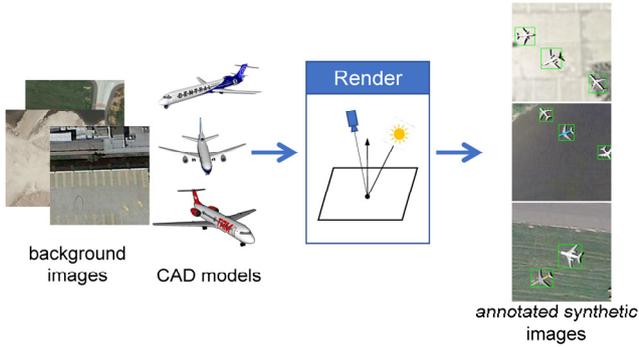

Fig. 1. Synthetic image generation by domain randomization. The 3D models are randomly placed on the background image (negative real image). By the use of various rendering factors, the system can generate a large volume of varied labeled data.

2) We eval the proposed method on NWPU VHR-10 [2], UCAS-AOD [3] and DIOR [4] datasets with Faster R-CNN [18] and R-FCN [19]. Results demonstrate that the proposed method is effective for augmenting insufficient real data.

## II. PROPOSED APPROACH

The proposed approach is realized in two stages. Firstly, 3D computer-aided design (CAD) models are collected and a virtual scene is built. Large-scale synthetic images with ground-truth bounding boxes are then generated based on domain randomization. The image style is then translated from the synthetic domain to the real domain. The object detector is first initialized using synthetic data and then the whole initialized network is trained on real data with smaller learning rate. We assumed that making use of synthetic data can overcome the small size of real data to some extent.

### A. Synthetic Image Generation Pipeline

Aiming at the diversity of real scenes, the method of domain randomization randomly sets the illumination, attitude, background texture, and other parameters of the rendering process. Fig. 1 shows the rendering framework. Specifically, in this study, we focused on generating aerial images containing airplanes. We therefore collected airplane models covering five categories (airliner, swept wing, jet, fanjet, propeller plane) from the online 3D Model Repository.[1] Firstly, the multiple airplane 3D CAD models were randomly placed in the virtual scene. A real image was then randomly selected as the background. We simulated global illumination via ambient light, which illuminated all the objects in the scene evenly and globally. Directional light was also applied to simulate a distant sun-like light source, so as to produce shadows. By adding the two light sources to the scene, the illumination and shadows of the airplanes were simulated. For the scale variation of the airplanes, the camera's attitude and position were randomly set. For the remote sensing image object detection task, we set up camera viewpoints to imitate an aerial view. In addition, we randomly added fog and multiple distractors [11] (geometric shapes) to the scene. synthetic imagery containing several aircraft and to calculate the bounding boxes. We implemented the 3D rendering system using three.js (a JavaScript 3D library).[2]

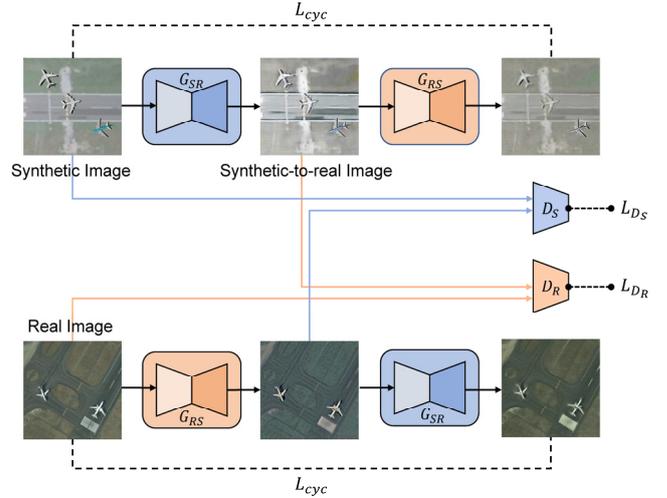

Fig. 2 Unpaired synthetic-to-real image translation using CycleGAN. The synthetic data domain is $S$, and the real data domain is $R$. The generator $G_{SR}$ maps the synthetic image to the real image, while the discriminator $D_S$ attempts to determine which domain the image comes from. Vice versa for $G_{RS}$ and $D_R$. The cycle consistency loss $L_{cyc}$ enforces simple forward-backward consistency.

### B. Unpaired Synthetic-to-Real Image Translation

In this study, we refined the synthetic images using all the positive images, without paired input-output examples. We assume that the CycleGAN technique does not change the geometry of the objects in the synthetic images, so that the annotations of the synthetic-to-real data can be inherited from the synthetic data.

The synthetic data domain is denoted as $S$, and the real data domain is denoted as $R$. The network architecture of CycleGAN is shown in Fig. 2. The generator $G_{S \to R}$ aims to map the synthetic image to the real image, while the discriminator $D_S$ tries to distinguish between the real domain and the synthetic domain. In contrast, the generator $G_{R \to S}$ maps the real image to the synthetic image, while the discriminator $D_R$ aims to distinguish which domain the sample comes from. The model is composed of two symmetric mappings. It learns not only the transfer from $S$ to $R$, but also from $R$ to $S$. The adversarial loss from synthetic data to real data is written as:

$$L_{adv_{S \to R}} = \min_{G_{S \to R}} \max_{D_R} \mathbb{E}_{r \sim p_{data}(r)}[\log D_R(r)] + \mathbb{E}_{s \sim p_{data}(s)}[\log(1 - D_R(G_{S \to R}(s)))] \quad (1)$$

where $s \sim p_{data}(s)$ is the synthetic data distribution and $r \sim p_{data}(r)$ is the real data distribution. The adversarial loss from real images to synthetic images is as follows:

$$L_{adv_{R \to S}} = \min_{G_{R \to S}} \max_{D_S} \mathbb{E}_{s \sim p_{data}(s)}[\log D_S(s)] + \mathbb{E}_{r \sim p_{data}(r)}[\log(1 - D_S(G_{R \to S}(r)))] \quad (2)$$

The loss of cycle consistency is introduced to keep the image consistent after translation and inverse translation. We thus assign the bounding box data of the synthetic image to the

---

[1] https://www.shapenet.org/

[2] https://threejs.org/

synthetic-to-real image directly. The objective is as follows:

$$L_{cyc} = \left\|G_{R \to S}\left(G_{S \to R}(s)\right) - s\right\|_1 + \left\|G_{S \to R}\left(G_{R \to S}(r)\right) - r\right\|_1 \quad (3)$$

Therefore, the full objective is:

$$L = L_{adv_{S \to R}} + L_{adv_{R \to S}} + \lambda L_{cyc} \quad (4)$$

where $\lambda$ is a weight for balancing the two objectives. We set $\lambda = 10$ in all the experiments empirically.

## III. EVALUATION

To evaluate the proposed approach in the object detection task, we compared the results of training on a small amount of real data, synthetic-to-real data, and a combination of the two. In this section, we first introduce the three datasets: NWPU VHR-10, UCAS-AOD and DIOR. We then describe the experimental details. Finally, we compare the visual differences between these real datasets and synthetic datasets.

### A. Dataset

NWPU VHR-10 is a 10-class dataset for remotely sensed image object detection. It contains a total of 800 VHR images. In total, the dataset contains 650 images containing at least one target, which are called positive images, while the rest are called negative images. As this work was focused on airplane detection, 90 images with 757 airplane instances were selected. The width of each image is about 1000 pixels. For the synthetic image generation, the negative real images were used as the background, i.e., water, buildings, trees, etc. For the synthetic-to-real style transfer, each of the 90 images was cropped to 256 × 256 pixels as the target domain. Since the CycleGAN model is a fully convolutional network, it can also infer larger images when testing. For airplane detection, 90 images were split randomly, with half for training and half for testing. We then cropped the optical images to 512 × 512 pixels, which was the same as the synthetic images.

UCAS-AOD is a 2-class dataset for high-resolution aerial vehicle and aircraft detection, made up of 2420 images from Google Earth. Only the plane subset of this dataset was selected, consisting of 7482 plane instances. As with NWPU VHR-10, we used the negative images as the background. For the synthetic-to-real style transfer, we cropped the positive images to 512 × 512 pixels. For the aircraft detection, 1000 positive images were randomly divided into a training set of 500 images and a test set of 500 images.

DIOR is a 20-class dataset for object detection in optical remote sensing images. We select all images of the airplane subset. There are 344 images in the training set and 705 images in the test set. The size of the image is 800 × 800 pixels.

### B. Experimental Details

CycleGAN: We adopted a TensorFlow implementation, as recommended by the author.[3] For $L_{adv_{S \to R}}$ and $L_{adv_{R \to S}}$, we used stable least-squares loss. We used the Adam optimizer with a learning rate of 0.0002 for the first 100k steps (~10 epochs), and a linear decay rate that went to zero over the next 100k steps. Note that all the real positive images without annotations were used as the target domain in the image translation experiment.

[3] https://github.com/vanhuyz/CycleGAN-TensorFlow

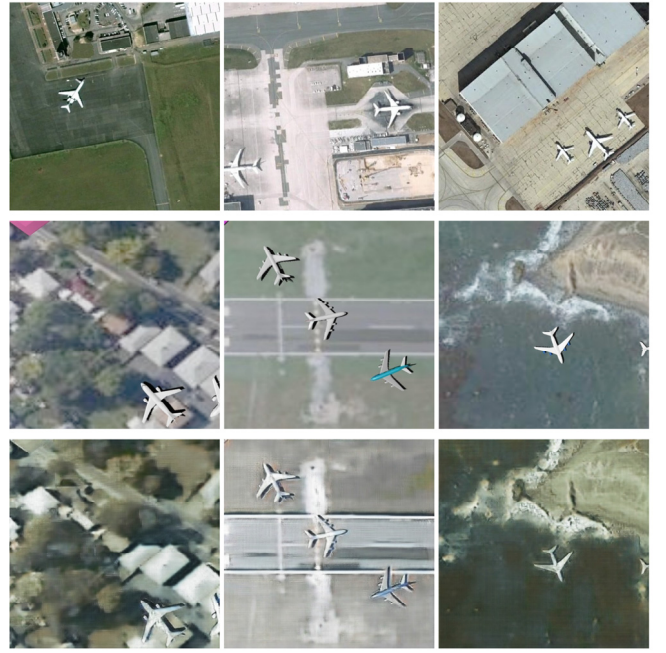

Fig. 3. Qualitative results for the synthetic images after translation. The real images are shown in the first row, the synthetic images are shown in the middle row, and the synthetic-to-real images are shown in the bottom row.

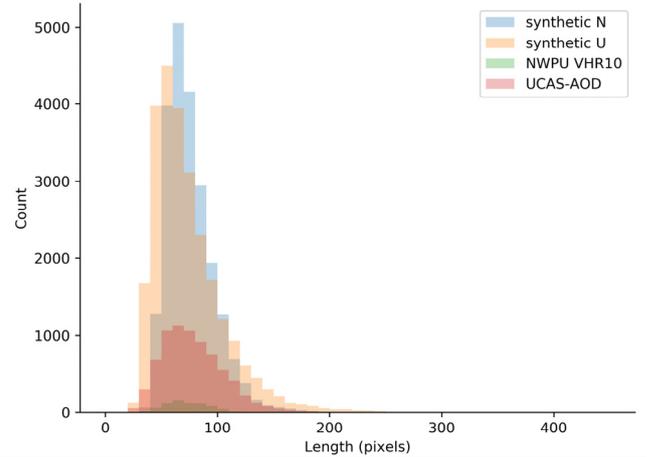

Fig. 4. Comparison between the synthetic data and real data in instance size. The synthetic data are Synthetic N (for NWPU VHR-10) and Synthetic U (for UCAS-AOD).

Faster R-CNN: The object detection code was based on the TensorFlow Object Detection API.[4] Resnet-101 [20] pre-trained on ImageNet [21] was used as the feature extractor. For the training, we used a momentum optimization algorithm with a value of 0.9 and a learning rate of 0.0003. The other parameters were set as the default values. The only data enhancement was horizontal flipping of the images. Each image is resized to 600 × 600 pixels. For the fine-tuning, the learning rate was reduced by 10 times, with the remaining hyperparameters unchanged. We used the early-stopping strategy to reduce over-fitting.

[4] https://github.com/tensorflow/models/tree/master/research/object_detection

R-FCN: Similar to Faster R-CNN, we select the momentum optimization algorithm with a value of 0.9, and the learning rate is 0.0003.

The experiments were carried out on a single GeForce RTX 2080Ti GPU with 12 GB of memory. For the results evaluation, we adopt the primary challenge metric of MSCOCO [22]. We report AP@0.75 and mAP@[0.5:0.05:0.95], respectively.

## C. Comparing Real, Synthetic, and Synthetic-to-Real Images

For the NWPU VHR-10 dataset, 10k synthetic images with ground-truth annotations were generated. Fig. 3 shows sample synthetic-to-real images, along with real and synthetic images. At the image level (e.g., style, illumination, appearance), the synthetic-to-real images are more realistic than the synthetic images. As the aircraft are compatible with the surrounding environment, the image translator makes the foreground and background more harmonious. We believe that this will help the detection model to focus on aircraft's essential features.

For the UCAS-AOD dataset, we further expanded the diversity of the synthetic images. Specifically, the number of background images was increased from 150 to 900, and the number of 3D CAD models was also increased from 19 to 30. These two synthetic datasets were termed Synthetic N (for NWPU VHR-10) and Synthetic U (for UCAS-AOD). Correspondingly, the translated datasets were called Synthetic-to-real N and Synthetic-to-real U.

One of the difficulties in remotely sensed image aircraft detection is the wide variety of sizes. Changes in the object size have important effects on the detection performance. The mean of the height and width of the bounding box is taken as a measurement of the instance size. Fig. 4 shows the length of the airplane instances in the synthetic images and real images. The synthetic dataset contains a larger number of instances, and those instances have a wider variety of sizes, which will be useful when generalizing the model.

## IV. RESULTS AND DISCUSSION

### A. Experimental Results for NWPU VHR-10

We first validated the proposed method on the NWPU VHR-10 dataset using the Synthetic-to-real N dataset. Real + synthetic means that the detection model first trains on the synthetic-to-real data, and then fine-tunes with the real data. As reported in Table I, the object detection network trained with the real training set of 50% real images achieves 64.10% mAP and 79.48% AP@0.75, which is used as the ideal case. The mAP of training on the 10k synthetic images is 45.00%, while Faster R-CNN achieves 60.50% mAP (close to the ideal case) when synthetic images are translated. This indicates that using style alignment to reduce the reality gap can improve the performance dramatically. For real tasks, although DR technology increases the diversity of the samples, the generalization ability of the models trained on synthetic images is not sufficient.

Next, we investigated whether a combination of the two datasets was required. The fine-tuning results with different numbers of real images are reported in Table I. By using all the real images for the fine-tuning, the mAP is increased by 4.36% (64.10% to 68.46%). The performance of training based on the pre-trained ResNet101 is lower than that of fine-tuning on the synthetic-to-real images. This demonstrates that the model trained on synthetic-to-real samples with a wide distribution is generalizable. Using synthetic-to-real data augmentation effectively avoids the problem of the lack of real data. The results of using R-FCN as the object detector is similar as that of Faster R-CNN, which also confirm our conclusion.

In addition, to explore the role of the number of synthetic images, we also conducted ablation experiments (Fig. 5). The model trained with synthetic-to-real images performs better than with synthetic images. With the increase in the number of synthetic images, the accuracy increases greatly. This shows that, due to the blindness of domain randomization, it depends on a large volume of data to achieve competitive results.

TABLE I
AIRCRAFT DETECTION RESULTS EVALUATED ON NWPU VHR-10 WITH DIFFERENT DOMAIN TRAINING DATA

| Training Data | Faster R-CNN | | R-FCN | |
|---|---|---|---|---|
| | mAP (%) | mAP@0.75 (%) | mAP (%) | mAP@0.75 (%) |
| Synthetic N 10k | 45.00 | 41.94 | 47.95 | 44.52 |
| Synthetic-to-real N 10k | 60.50 | 71.30 | 61.12 | 69.98 |
| 10% real | 54.10 | 53.96 | 53.86 | 51.41 |
| 10% real + syn | 62.09 (+ 7.99) | 76.25 (+ 22.29) | 59.52 (+5.66) | 71.44 (+20.03) |
| 20% real | 56.09 | 55.33 | 57.36 | 61.15 |
| 20% real + syn | 64.31 (+ 8.22) | 77.55 (+ 22.22) | 62.08 (+4.72) | 74.91 (+13.76) |
| 50% real | 64.10 | 79.48 | 63.15 | 76.15 |
| 50% real + syn | **68.46** (+ 4.36) | **86.52** (+ 7.04) | **66.29** (+3.14) | **82.72** (+6.57) |

TABLE II
AIRCRAFT DETECTION RESULTS EVALUATED ON UCAS-AOD WITH DIFFERENT DOMAIN TRAINING DATA

| Training Data | Faster R-CNN | | R-FCN | |
|---|---|---|---|---|
| | mAP (%) | mAP@0.75 (%) | mAP (%) | mAP@0.75 (%) |
| Synthetic U 10k | 27.89 | 15.04 | 29.32 | 16.46 |
| Synthetic-to-real U 10k | 38.91 | 20.91 | 38.96 | 24.53 |
| 1% real | 42.15 | 33.35 | 41.34 | 32.37 |
| 1% real + syn | 48.23 (+ 6.08) | 48.12 (+ 14.77) | 44.47 (+ 3.13) | 39.22 (+6.85) |
| 10% real | 62.53 | 76.15 | 62.12 | 75.93 |
| 10% real + syn | **64.87** (+ 2.34) | **80.72** (+ 4.57) | **62.33** (+ 0.21) | **76.85** (+0.92) |

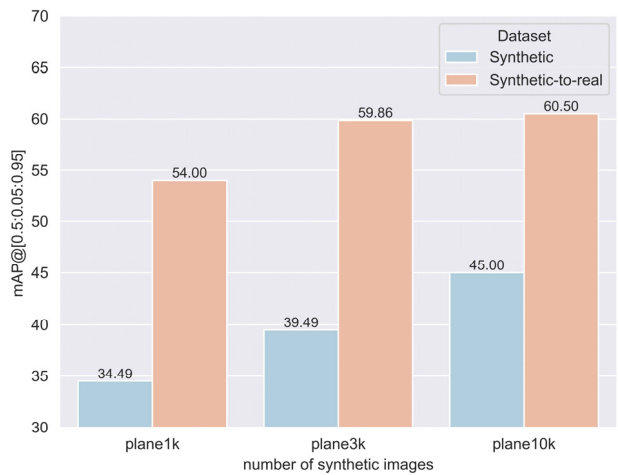

Fig. 5. Performance of Faster R-CNN as a function of the number of training images used, for both the synthetic data (blue) and the translated synthetic data (orange).

TABLE III
AIRCRAFT DETECTION RESULTS EVALUATED ON DIOR WITH DIFFERENT
DOMAIN TRAINING DATA

| Training Data | MAP (%) | AP@0.75 (%) |
| --- | --- | --- |
| Synthetic-to-real N 10k | 17.15 | 17.74 |
| Synthetic-to-real U 10k | 20.98 | 16.83 |
| 10% real | 15.80 | 13.51 |
| 10% real + Syn N | 19.45 (+3.65) | 20.08 (+6.57) |
| 10% real + Syn U | 20.00 (+4.20) | 19.66 (+6.15) |
| 20% real | 19.23 | 19.03 |
| 20% real + Syn N | 20.81 (+ 1.58) | 21.78 (+ 2.75) |
| 20% real + Syn U | 22.50 (+ 3.27) | 22.42 (+ 3.39) |
| 100% real | **32.73** | **33.02** |

*B. Experimental Results for UCAS-AOD*

We also tested the proposed approach with the UCAS-AOD dataset using the Synthetic-to-real U dataset. Again, the mAP for the synthetic-to-real images (38.91%) is higher than for the synthetic images (27.89%) (Table II), which again proves the need for image translation. Compared with ResNet101 pre-trained on ImageNet, the Faster R-CNN model fine-tuned on the synthetic samples generally achieves a higher mAP. By fine-tuning on 1% of real data, the proposed method achieves a performance of 48.23%, which is 6.08% higher than when using pure real data. The results show that the proposed method can effectively improve the detection accuracy in a real task, especially when there are few real data samples available. On the other hand, as the real data increases, the synthetic data become less important.

*C. Experiments Results for DIOR*

In order to eval the generalization of our proposed method, we apply synthetic-to-real data to DIOR dataset which has a large range of object size variations. As shown as Table III, the object detection network trained with all the real training set achieves 32.73% mAP and 33.02% AP@0.75 (upper-bound performance). Both Synthetic-to-real N and Synthetic-to-real U are helpful for improve the performance of Faster R-CNN object detector trained exclusively on limit real images. Compared with Synthetic-to-real N, instance size of Synthetic-to-real U are more widely distributed, which may be one of the reasons why Synthetic-to-real U works better.

## V. CONCLUSION

In this letter, we have proposed a practical framework for making better use of synthetic images in the aerial image airplane detection task. In the experiments, we generated a very large number of annotated synthetic images with a wide variety of aircraft instances. The experimental results indicated that aligning the style of synthetic and real images can benefit the performance. Using synthetic-to-real images is a promising approach for data augmentation, especially when only a few real data samples can be obtained. In the future, we will extend this method to more application scenarios.

REFERENCES

[1] L. Liu, W. Ouyang, X. Wang, P. Fieguth, J. Chen *et al*, "Deep learning for generic object detection: A survey," *International Journal of Computer Vision*, vol. 128, no. 2, pp. 261–318. 2020.
[2] G. Cheng, P. Zhou, and J. Han, "Learning rotation-invariant convolutional neural networks for object detection in VHR optical remote sensing images," *IEEE Transactions on Geoscience and Remote Sensing*, vol. 54, no. 12, pp. 7405–7415, 2016.
[3] H. Zhu, X. Chen, W. Dai, K. Fu, Q. Ye, and J. Jiao, "Orientation robust object detection in aerial images using deep convolutional neural network," in *IEEE International Conference on Image Processing*, 2015, pp. 3735–3739.
[4] K. Li, G. Wan, G. Cheng, L. Meng, and J. Han, "Object detection in optical remote sensing images: A survey and a new benchmark," *ISPRS Journal of Photogrammetry and Remote Sensing*, vol. 159, pp. 296–307, 2020.
[5] J. Han, D. Zhang, G. Cheng, L. Guo, and J. Ren, "Object Detection in Optical Remote Sensing Images Based on Weakly Supervised Learning and High-Level Feature Learning," *IEEE Transactions on Geoscience and Remote Sensing*, vol. 53, no. 6, pp. 3325–3337, 2015.
[6] G. Ros, L. Sellart, J. Materzynska, D. Vazquez, and A. M. Lopez, "The SYNTHIA dataset: A large collection of synthetic images for semantic segmentation of urban scenes," in *Proceedings of the IEEE conference on computer vision and pattern recognition*, 2016, pp. 3234–3243.
[7] S. R. Richter, V. Vineet, S. Roth, and V. Koltun, "Playing for data: Ground truth from computer games," in *European conference on computer vision*. Springer, 2016, pp. 102–118.
[8] X. Peng, B. Sun, K. Ali, and K. Saenko, "Learning deep object detectors from 3D models," in *Proceedings of the IEEE International Conference on Computer Vision*, 2015, pp. 1278–1286.
[9] H. A. Alhaija, S. K. Mustikovela, L. Mescheder, A. Geiger, and C. Rother, "Augmented reality meets computer vision: Efficient data generation for urban driving scenes," *International Journal of Computer Vision*, vol. 126, no. 9, pp. 961–972. 2018.
[10] X. Peng and K. Saenko, "Synthetic to Real Adaptation with Generative Correlation Alignment Networks," in *IEEE Winter Conference on Applications of Computer Vision*, 2018, pp. 1982–1991.
[11] J. Tremblay, A. Prakash, D. Acuna, M. Brophy, V. Jampani, C. Anil, T. To, E. Cameracci, S. Boochoon, and S. Birchfield, "Training deep networks with synthetic data: Bridging the reality gap by domain randomization," in *Proceedings of the IEEE Conference on Computer Vision and Pattern Recognition Workshops*, 2018, pp. 969–977.
[12] Y. Chen, W. Li, C. Sakaridis, D. Dai, and L. V. Gool, "Domain Adaptive Faster R-CNN for Object Detection in the Wild," in Proceedings of the IEEE Conference on Computer Vision and Pattern Recognition, 2018, pp. 3339–3348.
[13] Y. Yan, Y. Zhang, and N. Su, "A Novel Data Augmentation Method for Detection of Specific Aircraft in Remote Sensing RGB Images," *IEEE Access*, vol. 7, pp. 56051–56061, 2019.
[14] J. Tobin, R. Fong, A. Ray, J. Schneider, W. Zaremba, and P. Abbeel, "Domain randomization for transferring deep neural networks from simulation to the real world," in *IEEE/RSJ International Conference on Intelligent Robots and Systems*, 2017, pp. 23–30.
[15] G.-S. Xia, X. Bai, J. Ding, Z. Zhu, S. Belongie, J. Luo, M. Datcu, M. Pelillo, and L. Zhang, "DOTA: A large-scale dataset for object detection in aerial images," in *Proceedings of the IEEE Conference on Computer Vision and Pattern Recognition*, 2018, pp. 3974–3983.
[16] P. Isola, J.-Y. Zhu, T. Zhou, and A. A. Efros, "Image-to-image translation with conditional adversarial networks," in *Proceedings of the IEEE Conference on Computer Vision and Pattern Recognition*, 2017, pp. 1125–1134.
[17] J.-Y. Zhu, T. Park, P. Isola, and A. A. Efros, "Unpaired image-to-image translation using cycle-consistent adversarial networks," in *Proceedings of the IEEE International Conference on Computer Vision*, 2017, pp. 2223–2232.
[18] S. Ren, K. He, R. Girshick, and J. Sun. Faster R-CNN: Towards real-time object detection with region proposal networks. in *Proc. NIPS*, 2015, pp. 91–99.
[19] J. Dai, Y. Li, K. He, and J. Sun. R-FCN: Object detection via region-based fully convolutional networks. in *arXiv*:1605.06409, 2016.
[20] K. He, X. Zhang, S. Ren, and J. Sun, "Deep residual learning for image recognition," in *Proceedings of the IEEE Conference on Computer Vision and Pattern Recognition*, 2016, pp. 770–778.
[21] J. Deng, W. Dong, R. Socher, L.-J. Li, K. Li, and L. Fei-Fei, "Imagenet: A large-scale hierarchical image database," in *IEEE Conference on Computer Vision and Pattern Recognition*, 2009, pp. 248–255.
[22] T.-Y. Lin, M. Maire, S. Belongie, J. Hays, P. Perona, D. Ramanan *et al*, "Microsoft coco: Common objects in context," in *European conference on computer vision*. Springer, 2014, pp. 740–755.